\documentclass[runningheads,table]{llncs} 
\usepackage[table,dvipsnames,svgnames,x11names]{xcolor}
\usepackage{graphicx}

\usepackage{tikz}
\usepackage{comment}
\usepackage{amsmath,amssymb} 
\usepackage{color}
\usepackage{minitoc}
\usepackage{booktabs}
\usepackage{pifont}
\usepackage{makecell}
\usepackage{stfloats}
\usepackage{wrapfig}
\usepackage{appendix}
\usepackage{multirow}
\usepackage[normalem]{ulem}
\usepackage{enumitem}
\usepackage[pagebackref,breaklinks,colorlinks]{hyperref}
\usepackage[capitalize]{cleveref}
\usepackage{cite}

\crefname{section}{Sec.}{Secs.}
\Crefname{section}{Section}{Sections}
\Crefname{table}{Table}{Tables}
\crefname{table}{Tab.}{Tabs.}
\newcommand{\cmark}{\color{green}{\ding{51}}}%
\newcommand{\xmark}{\color{red}{\ding{55}}}%
\newcommand{\etal}{\textit{et al}.} 
\newcommand{\ie}{\textit{i}.\textit{e}. }
\newcommand{\eg}{\textit{e}.\textit{g}. }

\usepackage[accsupp]{axessibility}  

\definecolor{darkgreen}{RGB}{30,150,30}

\makeatletter
\renewcommand*{\@fnsymbol}[1]{\ensuremath{\ifcase#1\or *\or \dagger\or \ddagger\or
   \mathsection\or \mathparagraph\or \|\or **\or \dagger\dagger
   \or \ddagger\ddagger \else\@ctrerr\fi}}
\makeatother

\begin{document}
\pagestyle{headings}
\mainmatter
\def\ECCVSubNumber{2336}  

\title{Hierarchical Semantic Regularization of Latent Spaces in StyleGANs} 

\titlerunning{Hierarchical Semantic Regularization of Latent Spaces in StyleGANs}
%



\author{Tejan Karmali\inst{1,2}\thanks{Work done while at Indian Institute of Science} \and
Rishubh Parihar\inst{1} \and
Susmit Agrawal\inst{1}\and\\
Harsh Rangwani\inst{1} \and
Varun Jampani\inst{2}\and
Maneesh Singh\inst{3}\thanks{ Work done while at Verisk Analytics} \and\\
R. Venkatesh Babu\inst{1}}
\authorrunning{Karmali et al.}
%
\institute{Indian Institute of Science, Bengaluru, India 
\\ \and
Google Research \and Motive Technologies, Inc}
\maketitle

\begin{abstract}
Progress in GANs has enabled the generation of high-res\-olution photorealistic images of astonishing quality. StyleGANs allow for compelling attribute modification on such images via mathematical operations on the latent style vectors in the $\mathcal{W}/\mathcal{W+}$ space that effectively modulate the rich hierarchical representations of the generator. Such operations have recently been generalized beyond mere attribute swapping in the original StyleGAN paper to include interpolations. In spite of many significant improvements in StyleGANs, they are still seen to generate unnatural images. The quality of the generated images is predicated on two assumptions; (a) The richness of the hierarchical representations learnt by the generator, and, (b) The linearity and smoothness of the style spaces. In this work, we propose a Hierarchical Semantic Regularizer (HSR)\footnote[1]{Project Page and code:  \href{https://sites.google.com/view/hsr-eccv22}{https://sites.google.com/view/hsr-eccv22}}  which aligns the hierarchical representations learnt by the generator to corresponding powerful features learnt by pretrained networks on large amounts of data. HSR is shown to not only improve generator representations but also the linearity and smoothness of the latent style spaces, leading to the generation of more natural-looking style-edited images. To demonstrate improved linearity, we propose a novel metric - Attribute Linearity Score (ALS). A significant reduction in the generation of unnatural images is corroborated by improvement in the Perceptual Path Length (PPL) metric by $16.19\%$ averaged across different standard datasets while simultaneously improving the linearity of attribute-change in the attribute editing tasks. 

\end{abstract}

\section{Introduction}

\begin{figure*}[!t]
    \centering
    \includegraphics[width=\linewidth]{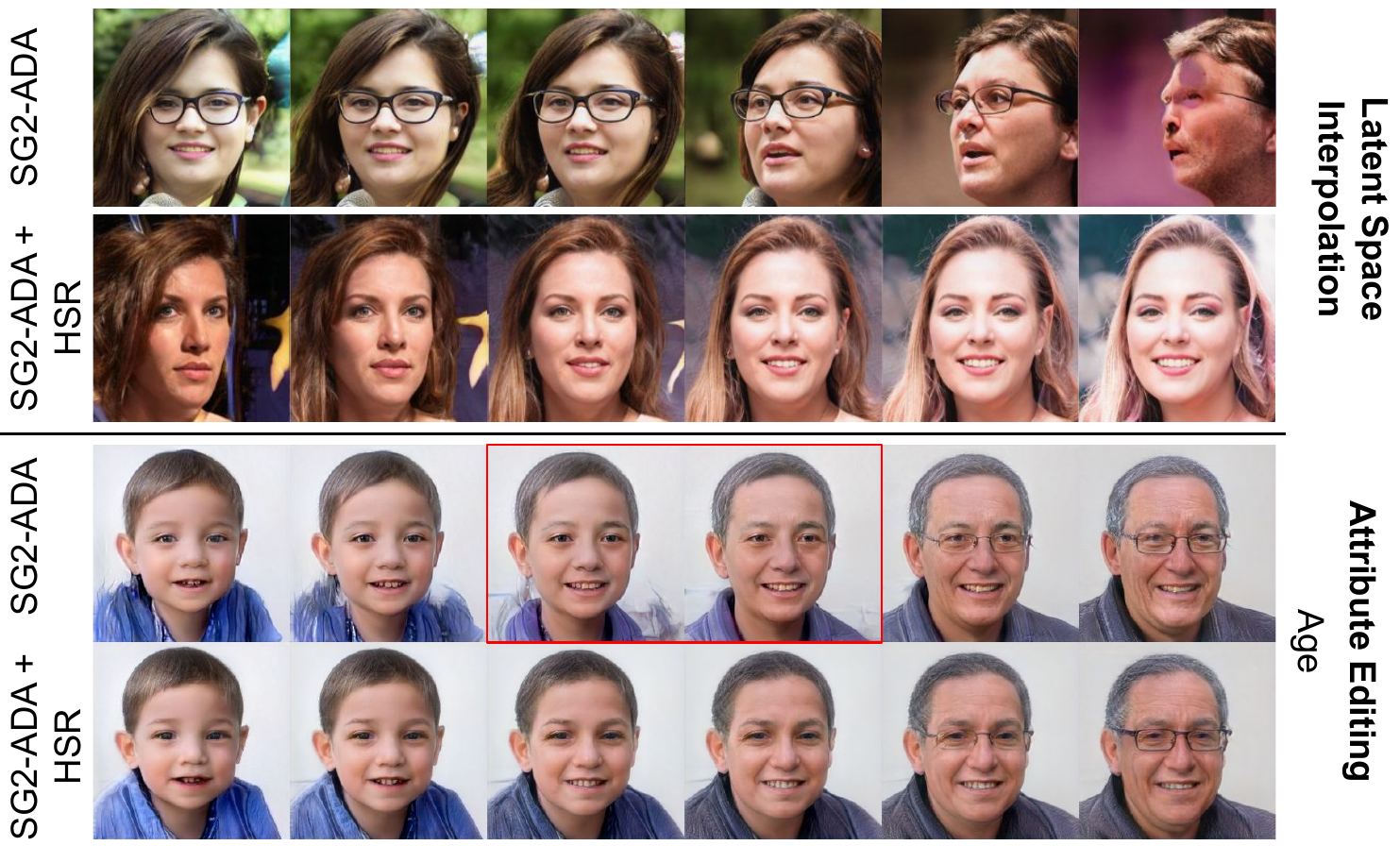}
    \caption{Hierarchical Semantic Regularizer (HSR) improves the latent space to semantic image mapping to produce more natural-looking images. \textbf{Top:} We show latent interpolation for images from bottom 10\%-ile image pairs ranked by PPL, a metric to measure smoothness of latent space. \textbf{Bottom:} Latent space using HSR mitigates artefacts in images during attribute edit transition and can transition smoothly (young to old (SG2-ADA) vs. young to middle-age to old (SG2-ADA+HSR), in continuous attributes like ``Age"). Zoom in to observe the effects.}
    \label{fig:teaser}
\end{figure*}

Recent years have seen tremendous advances in Generative Adversarial Network (GAN)~\cite{Goodfellow_GAN} architectures and their training methods to produce highly photorealistic images~\cite{biggan, Karras_2020_CVPR}.
Progress in the StyleGAN family of GAN architectures has shown promise by improving both the image quality, as well as the quality of latent space representations which enables controlled image generation. This is achieved by transforming an input noise space $\mathcal{Z}$ to a latent style space $\mathcal{W}$ which modulates a  synthesis network at various levels of representation hierarchies to generate an image with that style. This enables generation of compelling synthetic images with novel styles as well as practically useful applications such as GAN-based image attribute editing, style mixing, etc.~\cite{shen2020interfacegan,shen2021closedform,harkonen2020ganspace,Patashnik_2021_ICCV,patashnik2021styleclip,alaluf2021matter}. Nonetheless, such networks still often produce unrealistic images (ref. Fig.~\ref{fig:teaser}).

These quality issues in StyleGANs can have the following sources: (a) the hierarchical representation spaces in the synthesis network, (b) the latent style space, in particular the linearity and smoothness of such spaces, and (c) the functions used to transform the representation spaces in (a) using the corresponding hierarchical style vectors in (b). Our work seeks to address some of these issues.

We take inspiration from the recent advances in self-supervised and supervised learning~\cite{resnet,Caron_2021_ICCV,mocoimproved,vgg} which have allowed for the learning of semantically rich image representations translating into significant performance improves on image classification and other vision tasks~\cite{robusttransferimagenet,vgg,style_transfer}. Training on large datasets of natural images, like ImageNet~\cite{deng2009imagenet}, allows these techniques to learn hierarchically organized feature spaces capturing richer statistical patterns in natural images: shallower layer capturing low-level image features and the deeper layers abstract features highly correlated with visual semantics. Such pre-trained  representations can be harnessed to enhance the representational power of StyleGANs.

In fact, we demonstrate that transferring rich pretrained representations mentioned above allow us to mitigate simultaneously the challenges associated with both the representation spaces in the synthesis network as well as the latent style spaces modulating these representations. To allow for such a transfer, we propose to use a regularization mechanism, called the Hierarchical Semantic Regularizer (HSR) which aligns the generator's features to those from an appropriate, state of the art, pretrained feature extractor at several corresponding scales (levels) of the generator network. The architecture is shown in Fig.~\ref{fig:approach}. 

Karras \etal~\cite{karras2019style} introduced the Perceptual Path Length (PPL) metric to measure the smoothness of mapping from a latent space to the output image and showed its correlation with the generated image quality. We demonstrate that HSR regularization in StyleGAN training leads to $16.19\%$ relative improvement in PPL over StyleGAN2, leading to more realistic interpolations (refer Fig.~\ref{fig:teaser}). 

A standard approach for controlled synthesis of novel images is via linear (convex) interpolation between attributes\footnote{We use style and attributes interchangeably.} corresponding to real images. Applications such as image editing utilize such capabilities under the presumption that style spaces are both linear as well as decorrelated allowing for desired controlled edits. Since, PPL does not measure linearity, we propose a  novel metric, Attribute Linearity Score (ALS), to measure linearity in the attibute space. 
We demonstrate that HSR simultaneously improves linearity leading to smoother edits with significantly reduced editing artifacts (Fig.~\ref{fig:teaser}). A mean relative improvement of 15.5\% over StyleGAN2-ADA is achieved on the ALS metric.

Our contributions are: (a) A novel Hierarchical Semantic Regularizer (HSR) improving the generation of natural-looking synthetic images from StyleGANs. HSR is presented in (Sec.~\ref{sec:approach} with an analysis of design choices ~\ref{sec:design_disc}). (b) Extensive bench-marking of improvements by HSR  regularization on popular datasets, especially when utilizing linear interpolations for attribute editing (Sec.~\ref{sec:expts}). (c) Since linearity of the latent attribute space is very important for performing controlled edits, we propose a new metric, Attribute Linearity Score (ALS), in (Sec.~\ref{sec:latentspace}) and demonstrate improved linearity over the baselines.

\section{Related Works}
\label{sec:rel_works}
\noindent \textbf{Generative Adversarial Networks.} 
GAN proposed by Goodfellow \etal~\cite{Goodfellow_GAN}  a combination of two neural networks, \ie generator $G$ and discriminator $D$. For image synthesis the goal of $D$ is to differentiate between real and generated images, whereas the $G$ tries to fool the discriminator into classifying generated images as real. In the recent years several improvements in architecture~\cite{stylegan2,stylegan3, radford2015unsupervised, gulrajani2017improved, miyato2018cgans}, optimization objectives~\cite{albuquerque2019multi, arjovsky2017WGAN, mao2017least, mescheder2018R1}  and regularization~\cite{miyato2018spectral, gulrajani2017WGANGP} have made GANs an ubiquitous choice for image synthesis.  It has been observed that GANs developed for large scale datasets, suffer mode collapse when trained on limited data. Augmentation methods like DiffAugment~\cite{diffaug}, ADA~\cite{Karras2020ada}, ContraD~\cite{jeong2021contrad} , APA~\cite{jiang2021DeceiveD} etc. mitigate the collapse by reducing the discriminator's overfitting.

\noindent \textbf{Hierarchical Representations.} In classical vision, methods which decompose image into a hierarchy have been exploited for the tasks of image stitching, manipulation and fusion~\cite{adelson1984pyramid, choudhary2012pyramid}. Building on this motivation Shocher \etal \cite{shocher2020semantic} develop an image translation and manipulation method, which exploits hierarchical consistency of features of generator and a classifier. However this method is restricted to single image translation and manipulation. In contrast our work we aim to train a smooth and generalizable GAN which can simultaneously generate diverse images, by using semantic hierarchical consistency of features.

\noindent \textbf{Knowledge Transfer Using Pre-Trained Features.} 
Using pre-trained features trained on large scale datasets (\eg ImageNet etc.) \cite{he2016deep, tan2019efficientnet} have been useful for various downstream tasks across applications \cite{donahue2014decaf, huh2016makes, tzeng2017adversarial, yosinski2014transferable}. The recent development of the self-supervised approaches for representation learning~\cite{chen2020simple, radford2015unsupervised, grill2020bootstrap} have further immensely improved the quality of features learnt. These features are being used in various applications like part segmentation, localization etc. without being explicitly trained on such tasks~\cite{Caron_2021_ICCV}, which motivates our work which aims to transfer these semantic properties to $G$'s feature space. Currently much work for transfer learning for GANs has focused on the fine-tuning large GANs using a few images for adapting it to a different domain \cite{liu2020towards, mo2020FreezeD, ojha2021few, noguchi2019imagebsa}. Recently a concurrent work~\cite{kumari2021ensembling} also aims to use pre-trained features to improve GANs. However their goal is to improve discriminator. On contrary we aim to enrich GAN feature space by imparting it with semantic properties, leading to a disentangled and smooth latent space.

\noindent \textbf{Image Editing Using Latent Space Interpolations.} Latent space of pre-trained StyleGAN models is highly structured~\cite{shen2020interfacegan} and is popularly used to perform realistic image edits in the generated images~\cite{styleflow,shen2020interfacegan,wu2021stylespace,harkonen2020ganspace,shen2021closedform,latentclr,samage,parihar2022}. The primary idea in most of these approaches is to find a direction in the the extended latent space $\mathcal{W+}$ for editing attributes and transforming a latent code by moving in that direction to perform edits. StyleCLIP~\cite{patashnik2021styleclip} learns the directions for attribute editing by getting the guidance from pretrained CLIP~\cite{radford2021learning}. On the contrary, our work imposes constraints so that latent space has more naturally interpretable directions when used by the GAN-based image editing methods.


\section{Approach}
\label{sec:approach}
In this section, we first describe the objective of GAN framework, properties of StyleGAN, and its evaluation in Sec.~\ref{sec:bg}. Then, we describe Hierarchical Semantic Regularizer (HSR) (Sec.~\ref{sec:hsr}) and discuss its design in Sec.~\ref{sec:design_disc}.

\subsection{Preliminaries}
\label{sec:bg}
\noindent \textbf{Generative Adversarial Networks.} GAN involves two competing networks, namely a Generator $G$ and a Discriminator $D$. Taking a noise $\mathbf{z}$ sampled from a distribution $P_{z}$ as input, $G$ generates an image $G(\mathbf{z}) \in \mathbb{R}^{3 \times H\times W}$. Whereas, $D$ takes an input image ${x} \in \mathbb{R}^{3 \times H\times W}$, and tries to classify it as real or generated. The objective of $G$ is to fool $D$ into making it classify the generated image as a real one.
Formally, the learning objective can be written as:
\begin{equation}
\begin{split}
      &\max_D \mathcal{L}_{D} = \underset{x \sim P_{r}}{\mathbb{E}}[log(D(x))] + \underset{\mathbf{z} \sim P_{z}}{\mathbb{E}}[log(1 - D(G(\mathbf{z})))] \\
     & \min_G \mathcal{L}_{G} = \underset{\mathbf{z} \sim P_{z}}{\mathbb{E}}[log(1 - D(G(\mathbf{z})))] 
\end{split}
\end{equation}

\setlength{\intextsep}{0pt}%
\begin{wrapfigure}{r}{0.5\linewidth}
    \centering
    \includegraphics[width=\linewidth]{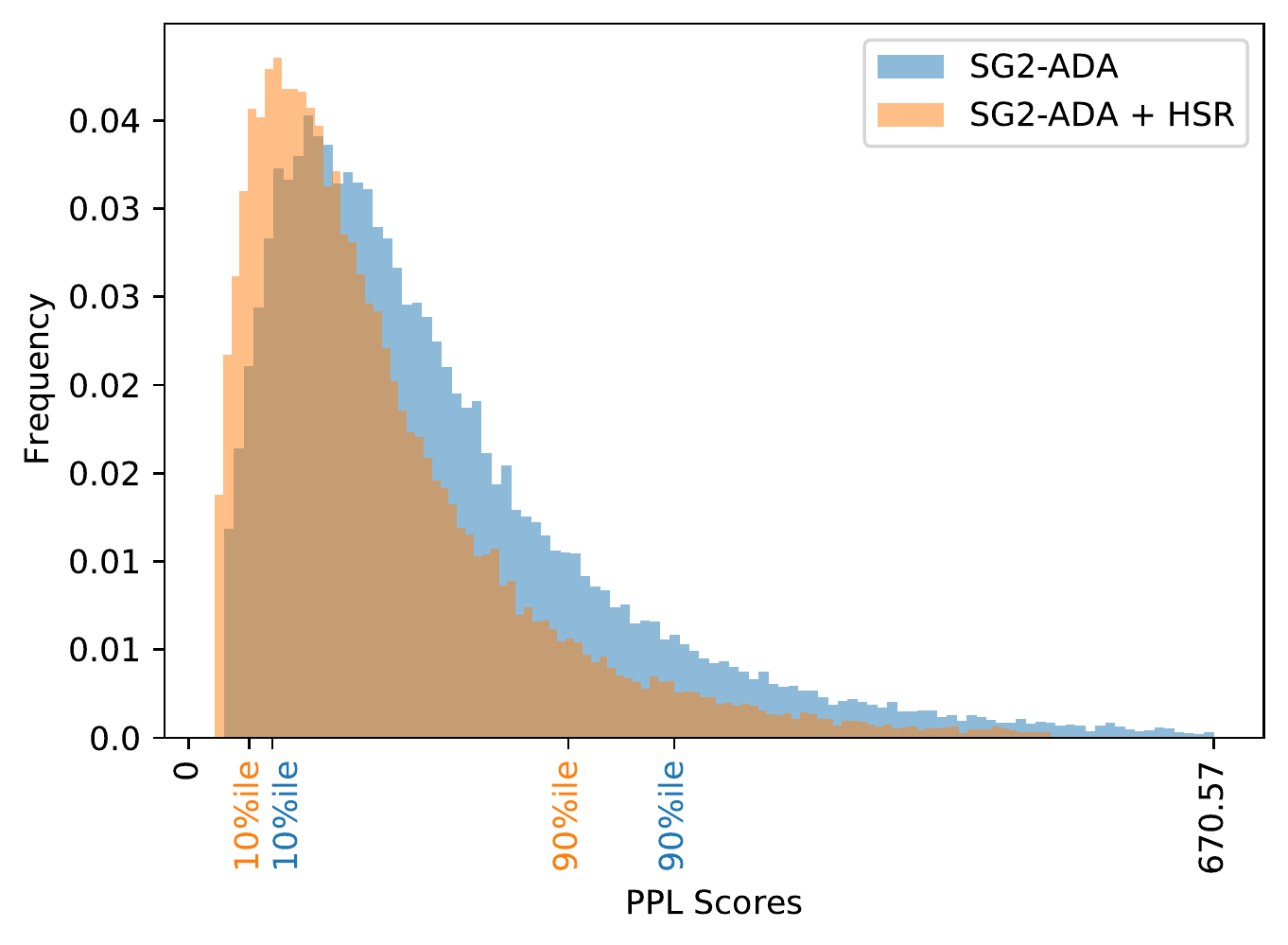}
    \caption{Distribution of PPL over 50k images from SG2-ADA and SG2-ADA+HSR. HSR improves the perceptual quality of top and bottom 10\%-ile images, thus leading to more natural-looking images.}
    \label{fig:ppl_hist}
\end{wrapfigure}

\noindent \textbf{StyleGAN.} In StyleGAN, an architectural modification is introduced where $\mathbf{z}$ is transformed into a semantic latent space through a sequence of linear layers called Mapping Network $G_m$, before generating the image $I$ through a Synthesis Network $G_s$ as $I = G_s(\mathbf{w})$. Hence, $G = G_s\circ G_m$.
The space learnt by $G_m$ is known as $\mathcal{W}+$-space. It is observed that $\mathcal{W}+$ is more meaningful in terms of attributes learned from the training data as compared to noise space $\mathcal{Z}$. 
Several methods~\cite{shen2020interfacegan,harkonen2020ganspace,shen2021closedform} propose ways to find attribute-specific directions in $\mathcal{W}+$ latent space.

\noindent \textbf{Perceptual Path Length.}
To measure the smoothness of the mapping from a latent space to the output image, Karras \etal ~\cite{Karras_2020_CVPR} proposed Perceptual Path Length (PPL). The requirement for this metric arises due to generation of unnatural images by GAN despite having low FID~\cite{Karras_2020_CVPR}. PPL aims to quantify the smoothness of latent space to output space mapping by measuring average of LPIPS~\cite{zhang2018unreasonable} distances between two generated images
under small perturbations in the latent space. A smoother latent space should have lesser PPL when compared to an uneven latent space.
It is shown~\cite{Karras_2020_CVPR} that PPL correlates well with image quality, \ie good quality images pairs will have less PPL, while if any one of the image is of bad quality, the PPL would be high. The images are sampled randomly without any truncation trick~\cite{kingma_glow,biggan} to compute PPL. As observed in Fig.~\ref{fig:qual_bottom}, the bottom 10\%-ile by PPL (sorted in increasing order) among the generated images appear as out-of-distribution images. Hence, the mean PPL score can be used to quantify the extent of non-smooth regions of latent space which produce unnatural images. Hence, we will be using this metric as a primary metric for comparison of the smoothness of latent space learnt by the models.

\subsection{Hierarchical Semantic Regularizer}
\label{sec:hsr}

\begin{figure}[t]
    \centering
    \includegraphics[width=0.95\linewidth]{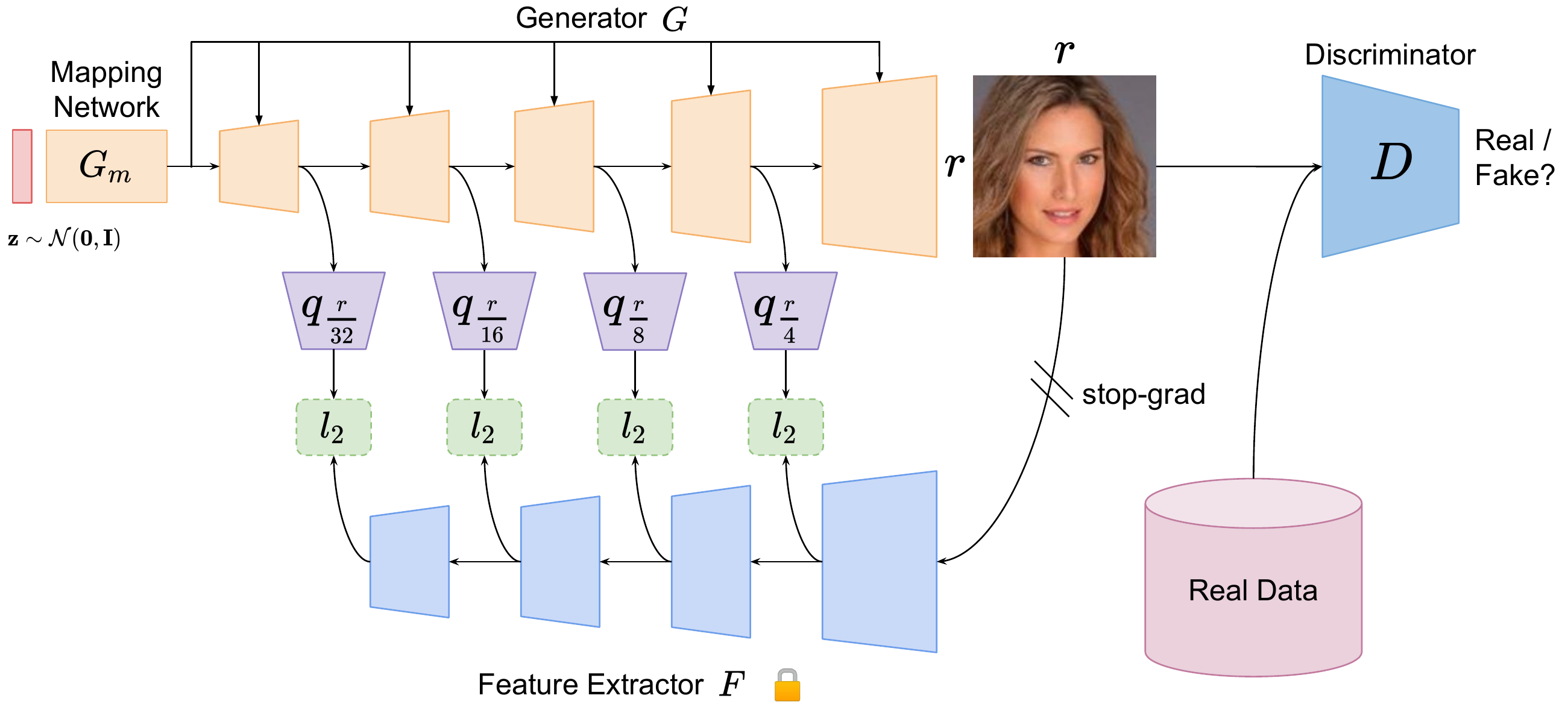}
    \caption{\textbf{Hierarchical Semantic Regularizer:} We use a pre-trained network to extract features at various resolution hierarchically. We then train linear predictors over generator features to predict the pre-trained features hierarchically. This transfers the semantic knowledge to generator feature space, making it's latent space meaningful, disentangled and editable. }
    \label{fig:approach_supp}
\end{figure}

Feature extractors of networks pretrained on large datasets (\eg ImageNet etc.) of natural images using classification or self-supervised losses store strong priors about the data, that are organized hierarchically. Each level of hierarchy captures a different semantic feature of data. The statistics of wide variety of natural images are captured by these networks~\cite{he2016deep,vitdino_seg, deit}. Due to the inherent differences in the nature of tasks, discriminative models capture different kinds of features compared to generative ones. Therefore, we seek to enrich the $G$'s intermediate feature space with guidance from a pretrained feature extractor. 

We first give a general idea of the proposed regularizer and then dive into various design choices made in it's formulation. Given an image $\mathbf{x}$ as input, the feature extractor $F$ returns semantically meaningful features from it. We attempt to make the generator aware of this explicit semantic feature space. To this end, we freeze the feature extractor and treat it as a fixed function that maps from image space to a semantically meaningful feature space.

Given such a mapping of the generated image, we try to align the Generator's features of this image through a set of feature predictors. This alignment is inspired by BYOL~\cite{grill2020bootstrap}.
As illustrated in Fig.~\ref{fig:approach_supp}, we attach a predictor branch $q$ to the Generator $G$. The objective of $q$ is to learn a mapping from generator's intermediate feature map $G^{\pi_G^i}(\mathbf{z})$ to pretrained feature extractor's intermediate feature $F^{\pi_F^i}(G(\mathbf{z}))$, where $\pi_G$ in $\pi_F$ denote the ordered set of layer numbers in the $G$ and $F$ at which we attach the predictors (ref. Eq.~\ref{eq:hsr}). We attach multiple such predictor networks $q_i$ at different scales of generator. 

\begin{equation}
\begin{split}
      \mathcal{L}_{G} = \underset{\mathbf{z} \sim P_{z}}{\mathbb{E}}[log(1 - D(G(\mathbf{z})))] + {\sum_{i=0}^{|\pi^i_G|}}\underset{\mathbf{z} \sim P_{z}}{\mathbb{E}}\|q(G^{\pi^i_G}(\mathbf{z})) - F^{\pi^i_F}(G(\mathbf{z}))\|_2^2
\end{split}
\label{eq:hsr}
\end{equation}

\subsection{Design Choices}
\label{sec:design_disc}
We analyse the effect of our Hierarchical Semantic Regularizer (HSR) against different design choices. For this purpose, we choose AnimalFace-Dog dataset which consists of 389 images. Since this is a low-shot dataset, we use StyleGAN2-ADA as our baseline. We perform all our experiments on $256\times 256$ resolution.

\noindent \textbf{What should be the choice of feature extractor?}
For this analysis, we choose 5 different feature extractors. We take combinations of CNN or transformer based networks trained using either self-supervised or supervised classification objective. We take ResNet-50 as the CNN backbone for both self-supervised (DINO) and supervised networks. For transformer-based networks, we use ViT-DINO and DeiT. Apart from trained networks, we also consider a randomly initialized ViT for baseline comparison.

We find that all pretrained feature extractors when used through HSR loss lead to introduction of meaningful semantic features in the intermediate latent spaces of the Generator. This is evidenced by reduction of PPL Score in Table~\ref{tab:feat_space}, which signifies reduction in non-meaningful generations from the GAN. The reduction in PPL also implies improved disentanglement~\cite{Karras_2020_CVPR} and linearity in the $\mathcal{W}$ space of the Generator, which is a desired property for many applications. We get $\ge 6.2\%$ improvement in the PPL score when guided by these networks.  ViT DINO's features stand apart, by improving the PPL score by ~19\% over the baseline. This is also supported by recent findings of Amir \etal ~\cite{vitdino_seg}, where they show several inherent properties of features from ViT-DINO, that are useful for computer vision tasks. With these results, we fix ViT DINO as the choice of the feature extractor for the rest of the experiments.

\noindent \textbf{Which layers of Generator are more important?}
The StyleGAN generator $G$ generates images using 7 synthesis blocks: starting from $4\times 4$, up to full resolution of $256 \times 256$. Of these, we consider synthesis blocks having features of resolution 8, 16, 32, 64. This corresponds to scaling down of resolution $r$ to $\frac{r}{32}$, $\frac{r}{16}$, $\frac{r}{8}$, and $\frac{r}{4}$. We choose these scales as it largely corresponds to the scales of downsampling by each block in SoTA CNN architectures~\cite{resnet,vgg}. The first block of $G$ (which have low resolution, but are responsible for high-level semantics) are supervised by the last block of the feature extractor (as they also are responsible for high-level semantics). Similarly, other blocks of $G$ are supervised by the blocks of the feature extractor that bring out similar level of semantics. 

To decide which layers contribute the most to the improvement in PPL, we divide the 4 blocks into 3 groups. The 3 groups specialize in high ($\frac{r}{32}$, $\frac{r}{16}$), mid ($\frac{r}{16}$, $\frac{r}{8}$), and low ($\frac{r}{8}$, $\frac{r}{4}$) level of semantics. We observe, in Table~\ref{tab:semantics}, that it is the supervision at low-level semantics which is most useful for the $G$. We observe a gradation in the improvement over the baseline, as high-level semantic supervision is least useful, followed by middle, and low. Overall, supervision at all levels turns out to cause the highest improvement.
\begin{table*}[t]
\parbox{.45\linewidth}{

\caption{\textbf{Feature space ablation}: Ablating over different feature extractors for usage in HSR. HSR with ViT-DINO's features gives best results.}
\label{tab:feat_space}
\begin{tabular}{lll}
\toprule
      {}          & {FID $\downarrow$}   & {PPL$\downarrow$}   \\ \midrule
StyleGAN2-ADA  & {53.28} & {59.27} \\ \hline
+ ViT (RandInit)  & {53.65}      & {56.97}  \\
+ ResNet50 DINO~\cite{Caron_2021_ICCV} & {54.33}        & 55.6     \\
+ DeiT~\cite{deit}         & 53.22          & 54.71    \\
+ ResNet50~\cite{resnet}      & 52.88          & 52.23    \\
+ ViT DINO~\cite{Caron_2021_ICCV}      & \textbf{51.58} & \textbf{48.02}        \\ \bottomrule
\end{tabular}
}
\hfill
\parbox{.55\linewidth}{
\centering
\caption{\textbf{Level of semantics}: A gradation in the improvement over the baseline is observed as we supervise from high-level semantics to low-level semantics. Best results are obtained when all the levels are supervised.}
\label{tab:semantics}
\begin{tabular}{lll}
\toprule
    {}     & {FID$\downarrow$} & {PPL$\downarrow$} \\ \midrule
StyleGAN2-ADA & 53.28                   & 59.27                   \\ \hline
+  High-level ($\frac{r}{32}$, $\frac{r}{16}$)  & 53.15                   & 57.73                   \\
+ Mid-level ($\frac{r}{16}$, $\frac{r}{8}$) & 52.91                   & 54.18                   \\
+ Low-level ($\frac{r}{8}$, $\frac{r}{4}$) & 53.66                   & 51.77                   \\ \hline
+    All levels    & \textbf{51.58}          & \textbf{48.02}          \\ \bottomrule
\end{tabular}
}
\end{table*}

\noindent \textbf{Does Path Length Regularizer (PLR) complement HSR?} Path Length Regularizer (PLR) was introduced in StyleGAN2~\cite{Karras_2020_CVPR}. The intuition behind PLR is to promote fixed magnitude non-zero change in the resulting image when moving by a fixed step size in the $\mathcal{W}+$-space. 
As reported in Table~\ref{tab:reg_abl}, we find that HSR itself gives slightly better improvement than the PLR over the baseline. While the best effect is noted when both, PLR and HSR, are applied together.
\noindent \textbf{Insight.} PLR's objective is to improve latent space smoothness, which leads to better PPL. Since PPL and image quality (natural-ness of image) are correlated, applying PLR improves the image quality. Whereas in HSR, we enforce the generator to predict in a feature space learnt from natural images using a pretrained feature extractor as prior. We observe that this objective, which targets bringing feature space of generator closer to a ``natural" feature space also leads to improvement in the smoothness of latent space, as measured by PPL. This shows that image quality and latent space smoothness are complementary and related concepts. Therefore, optimizing for both gives better PPL score. 
\section{Experiments}
\setlength{\intextsep}{0pt}
\begin{table*}[t]
\parbox{0.6\linewidth}{
\caption{\textbf{FFHQ-140k Results}: We report FID, Precision, Recall and PPL for different methods. With large data our method (SG2+HSR) produces better results even compared to SG2-ADA. }
\label{tab:full_data}
\resizebox{\linewidth}{!}{
\begin{tabular}{lllll}
\toprule
       FFHQ-140k    & FID$\downarrow$                                & Precision$\uparrow$                         & Recall$\uparrow$                              & PPL$\downarrow$                                 \\ \midrule
SG2             & {3.92}          & \textbf{0.68} & {0.45}          & {175.09}          \\
+ HSR    & \textbf{3.74} & \textbf{0.68} & \textbf{0.48} & \textbf{144.59} \\ \hline
SG2-ADA         & \textbf{4.30}                                   & 0.69                                   & \textbf{0.40}                                   & 163.11                                    \\

+ HSR    & 5.26 & \textbf{0.70} & 0.38 &  \textbf{131.41}\\ 
\bottomrule
\end{tabular}
}}
\parbox{0.4\linewidth}{
\centering
\caption{\textbf{Performance wrt PLR.} PLR and HSR complement each other, while being equally effective individually.}

\label{tab:reg_abl}
\begin{tabular}{cc|ll}

\toprule
PLR & HSR & {FID$\downarrow$} & {PPL$\downarrow$} \\ \midrule
\xmark & \xmark &   57.97   & 75.63                   \\
\cmark & \xmark &   53.28   & 59.27                   \\
\xmark & \cmark &   52.98   & 58.60                    \\
\cmark & \cmark & \textbf{51.58} & \textbf{48.02}          \\ \bottomrule
\end{tabular}

}

\end{table*}

\label{sec:expts}
In this section, we demonstrate the effectiveness of HSR experimentally. We first describe the experimental setup for all our experiments. Then, we evaluate the quantitative performance on several real-world datasets of varying sizes. Finally, we show improved linearity of latent space through attribute editing.
\subsection{Experimental Setup}
\noindent \textbf{Datasets.} We run our experiments on FFHQ~\cite{karras2018progressive} (70k images), AnimalFace-Dog (389 images), AnimalFace-Cat (160 images) ~\cite{si2011learning}, CUB200 (12k images) ~\cite{WahCUB_200_2011}, and LSUN-church ~\cite{yu15lsun} (126k images) (ref. supl. mat.) datasets. We augment the datasets by taking the horizontal flip of every image, doubling the number of images in the original dataset.
\noindent \textbf{Implementation Details.} We use StyleGAN2-ADA (SG2-ADA) as the baseline GAN, with its architecture for $256\times 256$ images, with batch size of 16. Predictors $q$ contain \texttt{Conv1x1-LeakyReLU-Conv1x1}, with hidden dimension of 4096. We make use of 2 A6000 GPUs for training our models.

\subsection{Results}
\begin{figure}
    \centering 
    \includegraphics[width=0.98\linewidth]{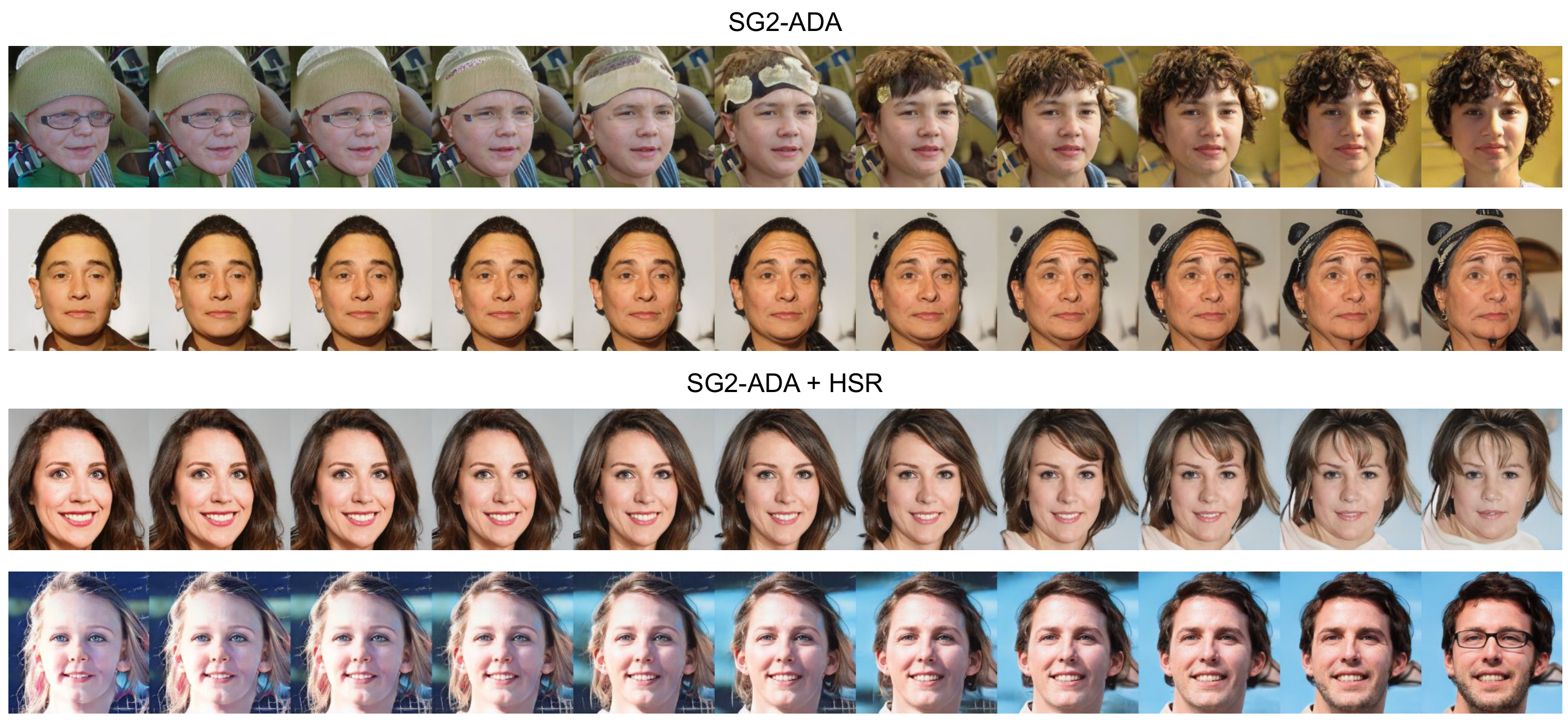}
    \caption{Latent space interpolation of top 10-\%ile images, ranked by PPL score. SG2-ADA images show traces of artifacts which are absent 
    after applying HSR.} 
    \label{fig:qual_top} 
\end{figure}

\begin{figure}[t]
    \centering
    \includegraphics[width=1.0\linewidth]{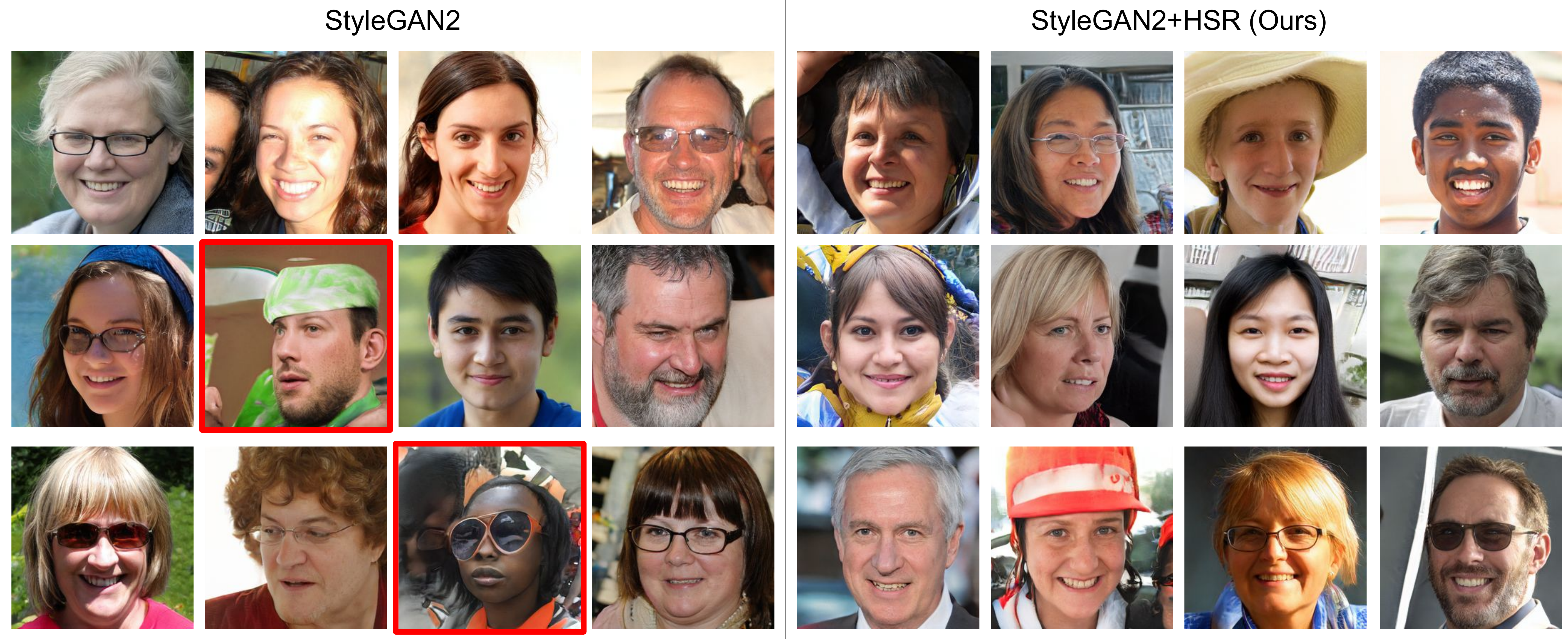}
    \caption{Comparison over uniformly random sampled images from StyeGAN2 and StyleGAN2+HSR. StyleGAN2 produces unnatural faces and artefacts over it such as peculiar eyeglasses (as shown in the highlighted images).}
    \label{fig:rand_sample}
\end{figure}

\begin{figure}[t]
    \centering 
    \includegraphics[width=0.98\linewidth]{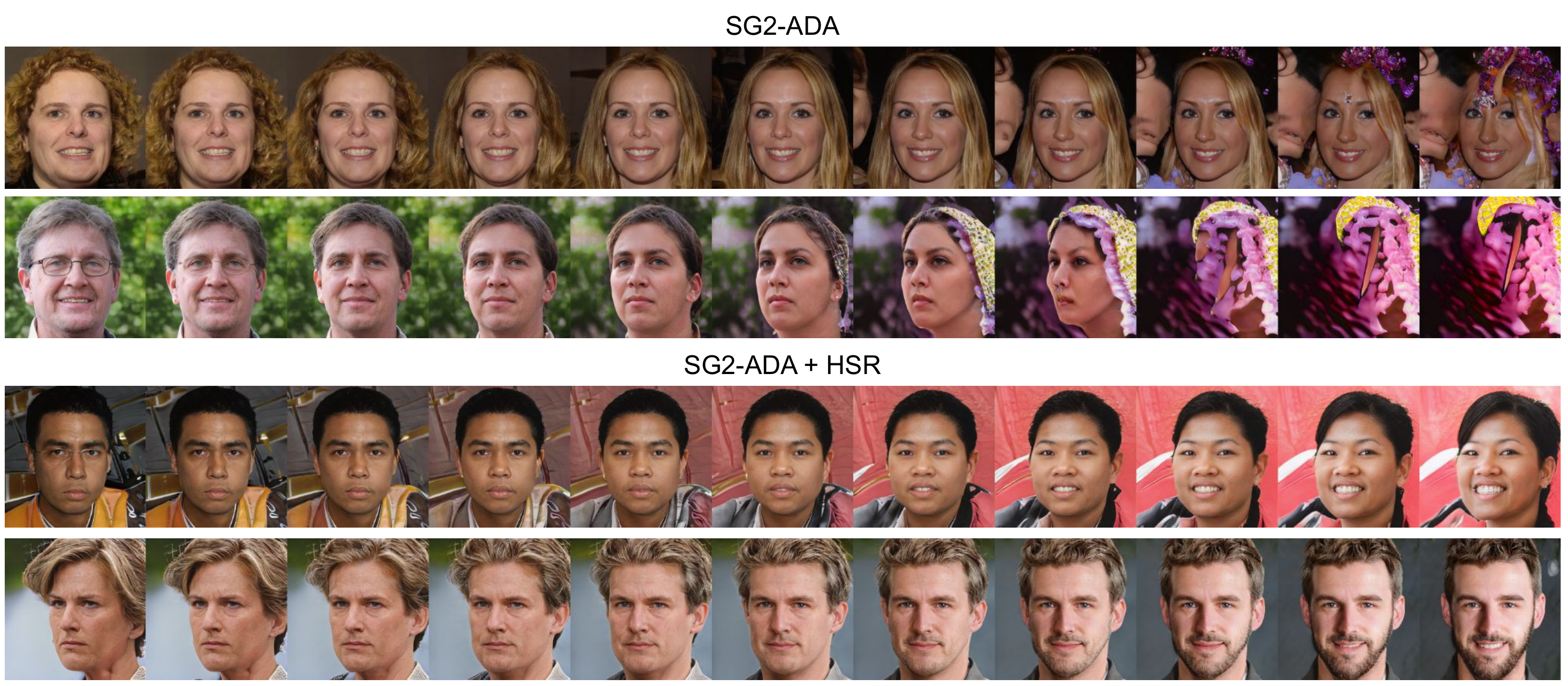}
    \caption{Latent space interpolation of bottom 10-\%ile images, ranked by PPL score. SG2-ADA latent space accommodates more unnatural images, leading to increase in PPL score. The latent space maps to more natural face-like images in SG2-ADA-HSR.} 
    \label{fig:qual_bottom} 
\end{figure}

On standard full dataset of FFHQ, we compare over both StyleGAN2~\cite{Karras_2020_CVPR} (SG2) and StyleGAN2-ADA~\cite{Karras2020ada}. This is because StyleGAN2 shows slightly better performance against the ADA variant on large datasets. 
We also evaluate our method for limited data sizes. Traditionally, GANs have shown to perform poorly on smaller datasets, until recently several approaches~\cite{Karras2020ada,jiang2021DeceiveD,yang2021dataefficient} have been proposed which enables GANs to learn well on limited data. We observe that irrespective of dataset size, asking the generator to be predictive of semantic features of rich feature extractors via HSR improves the smoothness of the latent space, as it is evident by an average relative improvement in PPL scores of about ~14.2\% on average in Table~\ref{tab:lim_data_res}, while that of 17.42\% in case of full FFHQ. This is also evident qualitatively in Fig.~\ref{fig:qual_top} and ~\ref{fig:qual_bottom}, where we observe an improved latent-to-image mapping even in bottom 10\%-ile images, when ranked by PPL scores. We also present images sampled randomly in Fig.~\ref{fig:rand_sample}, where we observe the mitigation of the unnatural faces and artefacts (highlighted images) upon application of HSR. Thus, HSR raises the lower bound for the natural-ness of the images produced by a generator (also ref. Fig.~\ref{fig:ppl_hist}).

\begin{table*}[t]
\caption{\textbf{Results on Limited Data} We present results on different limited data cases for FFHQ (left) dataset and on real-world datasets (right). We apply our regularizer on the strong baseline of StyleGAN2+ADA which is designed for limited data. We observe a significant decrease in PPL over baselines which implies a smooth, disentangled and meaningful latent space, while preserving photorealism (comparable FID). }
\label{tab:lim_data_res}
\parbox{0.49\linewidth}{
\resizebox{\linewidth}{!}{\begin{tabular}{llll}
\toprule
Dataset        & Method         & {FID$\downarrow$} & {PPL$\downarrow$}   \\ \midrule
\multirow{ 2}{*}{FFHQ-1k}       & StyleGAN2-ADA & \textbf{19.14}           & 98.79                    \\
               & + HSR         & 21.76                    & \textbf{90.39}           \\ \midrule
\multirow{ 2}{*}{FFHQ-2k}       & StyleGAN2-ADA & \textbf{14.74}           & 136.14                   \\ 
               & + HSR         & 15.53                    & \textbf{115.38}           \\ \midrule
\multirow{ 2}{*}{FFHQ-10k}       & StyleGAN2-ADA & \textbf{7.16}           & 164.21                    \\
               & + HSR         & 8.08                    & \textbf{126.35}           \\ \bottomrule
\end{tabular}}}
\parbox{0.51\linewidth}{
\resizebox{\linewidth}{!}{\begin{tabular}{llll}
\toprule
Dataset        & Method         & {FID$\downarrow$} & {PPL$\downarrow$}   \\ \midrule
{AnimalFace} & StyleGAN2-ADA & 53.28                   & 59.27                     \\
               Dog & + HSR         & \textbf{51.58}          & \textbf{48.02}            \\ \midrule
{AnimalFace } & StyleGAN2-ADA & \textbf{39.50}          & {50.76} \\
            Cat   & + HSR         & 40.25                   & \textbf{40.75}                     \\ \midrule
\multirow{ 2}{*}{CUB}             & StyleGAN2-ADA & {\textbf{5.78}}    & {265.46}      \\
               & + HSR         & {6.15}    & {\textbf{237.81}}      \\ \bottomrule
\end{tabular}}
}
\end{table*}

\subsection{Analysis of Linearity of Latent Space} 
\label{sec:latentspace}
\noindent \textbf{Motivation.}
Latent space of a pre-trained StyleGAN has meaningful directions embedded in it. Shen \etal ~\cite{shen2020interfacegan} shows that $\mathcal{W}+$ latent space is disentangled with respect to image semantics and there exist linear directions $\mathbf{d}$ in this space that control specific semantic attributes in the generated images. This is an important property of the latent space which is commonly used in controlled image synthesis~\cite{patashnik2021styleclip} and image editing ~\cite{styleflow}, as it leads to smooth interpolation between any two generated images. 
Furthermore, it is observed that the magnitude of latent transformations 
linearly correlates with the magnitude of the attribute changes in generated images~\cite{zhao2021large}.
Although, multiple works~\cite{styleflow,harkonen2020ganspace,shen2020interfacegan,wu2021stylespace,patashnik2021styleclip} are built upon this property to generate desired image transformations, there is no established metric
to evaluate the extent of this linear correlation in the latent space. To this end, we propose a new metric called \textit{Attribute Linearity Score (ALS)} for quantifying this linear correlation between the extent of latent transformations and the attribute changes. 

\begin{figure} 
    \centering
    \includegraphics[width=\linewidth]{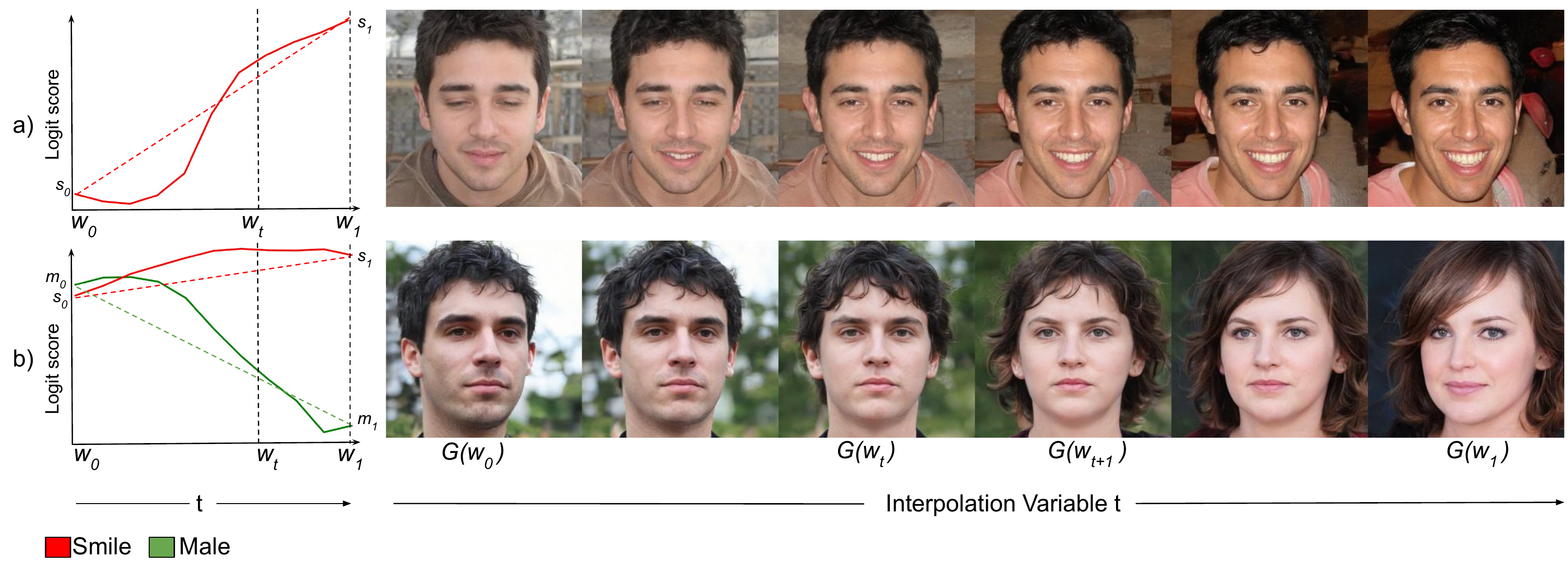}
    \caption{\textbf{Linearity of the latent space:} Here we show the transition images generated by the intermediate latent code $\mathbf{w_t}$ in the right and the corresponding attribute scores $s_t$ for for smile (row 1 and 2) and $m_t$ for male attribute (row 2). For brevity we have written $s_t=C_s(G(\mathbf{w_t}))$ and $m_t=C_m(G(\mathbf{w_t}))$.}  
    \label{fig:attr-visual-result}
\end{figure}

\noindent \textbf{Attribute Linearity Score (ALS).} Let the attribute strength be given by attribute score (logit value) 
from a pre\-trained attribute classifier $C$~\cite{karras2019style}. Consider two latent codes $\mathbf{w_0}$ and $\mathbf{w_1} \in \mathcal{W}+$ and their corresponding generated images $G(\mathbf{w_0})$ and $G(\mathbf{w_1})$ (using the generator $G$). Convex combinations of $\mathbf{w_0}$ and $\mathbf{w_1}$ generate interpolated latent codes  $\mathbf{w_t}$ (Eq.~\ref{eq:w-interpolate}) on the line segment joining the two latent codes $\mathbf{w_0}$ and $\mathbf{w_1}$. Let the corresponding generated images be denoted by $G(\mathbf{w_t})$. Linearity of the latent space ~\cite{zhao2021large} with respect to the attribute strength C implies that the attribute score for the image $G(\mathbf{w_t})$ should be the same convex combination of the attribute strengths of $G(\mathbf{w_0})$ and $G(\mathbf{w_1})$ (Eq. ~\ref{eq:score-interpolate}).
\begin{equation}
\label{eq:w-interpolate}
     \mathbf{w_t} = \mathbf{w_0} + t*(\mathbf{w_1} - \mathbf{w_0}),  \; t \in (0,1)
\end{equation}
\begin{equation}
\label{eq:score-interpolate}
C(G(\mathbf{w_t})) \approx C(G(\mathbf{w_0})) + t*(C(G(\mathbf{w_1})) - C(G(\mathbf{w_0}))),  \; t \in (0,1)
\end{equation}
Consider the example shown in Fig. ~\ref{fig:attr-visual-result}\textcolor{red}{a},
where we depict the transformation of the smile attribute. On the left, we show the plot of attribute scores with the interpolation parameter $t$ using smile classifier $C_s$ and on the right we show the image samples $G(\mathbf{w_t})$ for $t\in(0,1)$. A model with a linear latent space structure should have this plot close to the ``ideal" (shown in dotted) straight line between the two end points. Similar plots are shown for the ``smile" and ``male" attributes in Fig. ~\ref{fig:attr-visual-result}\textcolor{red}{b}. In both cases, we observe a significant departure from linearity.

\setlength{\intextsep}{0pt}



The ALS score quantifies the deviation from the line segment defined in Eq.~\ref{eq:score-interpolate} using the mean squared error metric. To compute this, we first define a set of equally-spaced interpolation points  $ t \in \{0, \frac{1}{N}, \frac{2}{N}, \dots, 1\}$. For each attribute $j \in \{1, \dots,M\}$), the squared difference ($\Delta_{tj}$) is computed using Eq. ~\ref{eq:3}. The ALS score ($\Delta_{T}$) is defined as the mean of $\Delta_{tj}$ over all $M$ attributes and $N$ interpolation points (\ie $ \Delta_{T} = \frac{1}{NM}\sum_{t=1}^{N}\sum_{j=1}^{M} \Delta_{tj}$).



\begin{equation}
\label{eq:3}
    \Delta_{tj} = ||C_j(G(\mathbf{w_t})) - C_j(G(\mathbf{w_0})) - t*(C_j(G(\mathbf{w_1}))-C_j(G(\mathbf{w_0})))||^2
\end{equation}


\begin{figure*}[t]
  \centering
\begin{minipage}[c]{0.58\linewidth}
\centering
\caption{ALS score comparison upon adding HSR. \textit{(Right)}: Mean ALS computed for each value of the interpolation variable \textit{t}. HSR is able to achieve a lower value of ALS supporting the linearity induced by ALS. \textit{(Bottom)}: ALS score computed for all the face attributes separately.} 
\label{tab:attribute_ppl_score} 
\resizebox{\textwidth}{!}{\begin{tabular}{lllllll|l} \\
\toprule
        &  Gender  & Smile & Age & Hair  & Bangs & Beard & Mean  \\ 
\midrule
SG2-ADA & 1.38 & 1.48 & 1.18 & 1.96 & 1.95   & 1.60 & 1.59 \\
+HSR & \textbf{1.12} & \textbf{0.99} & \textbf{1.15} & \textbf{1.87} & \textbf{1.62}  & \textbf{1.16} & \textbf{1.32}\\
\bottomrule
\end{tabular}
}
  \end{minipage}
  \hfill
  \begin{minipage}[c]{0.40\textwidth}
    \centering
    \includegraphics[width=\textwidth]{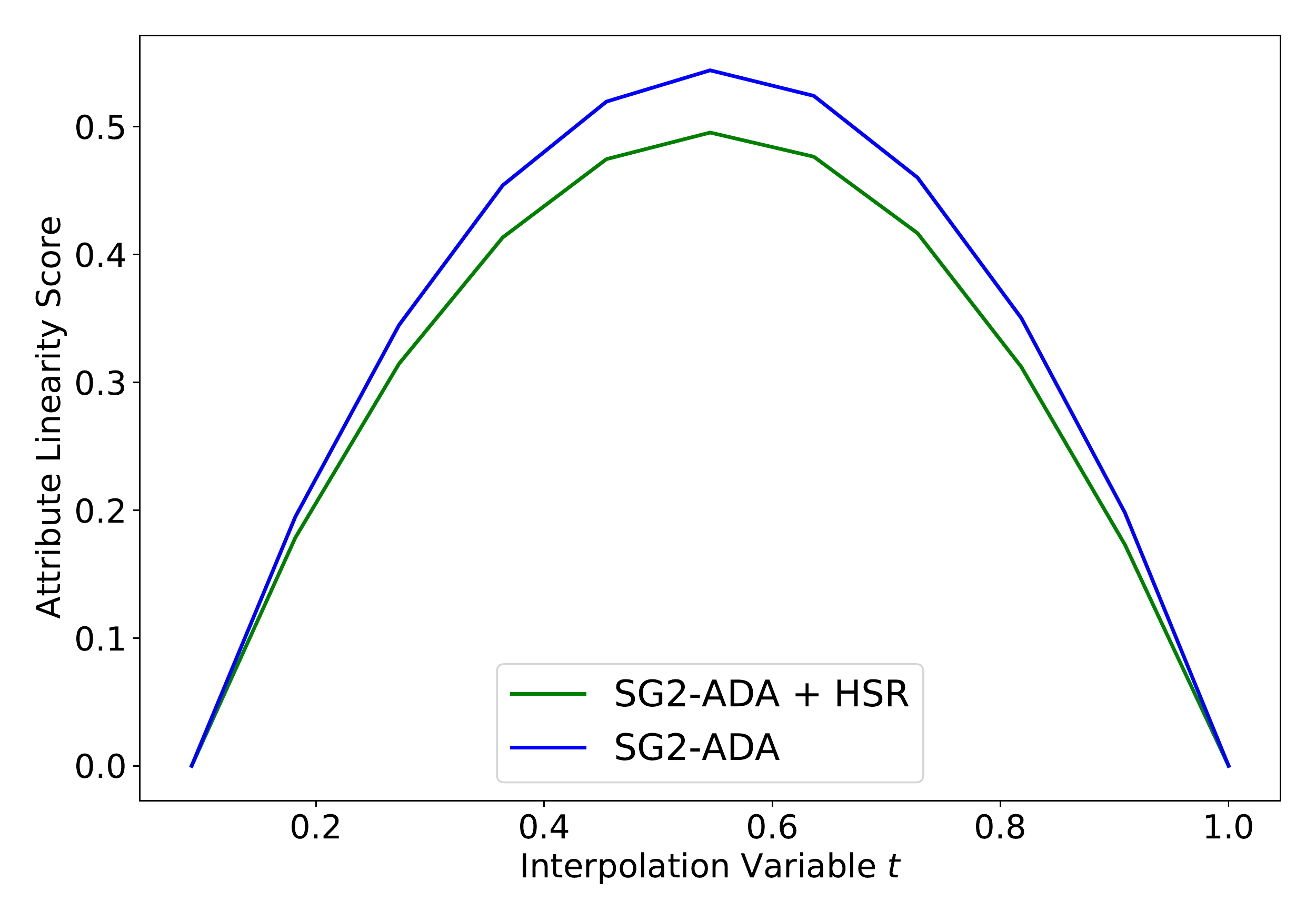}
    \begin{minipage}{1cm}
    \vfill
    \end{minipage}
    
    \label{fig:mean}
  \end{minipage}
\end{figure*}

In the following sections, we first evaluate effect of linearity on appplying HSR, by measuring ALS. Then we show it's application in measuring edits in images. We use StyleGAN2-ADA model as the baseline trained on FFHQ-10k for results in the rest of this section.
    
\noindent \textbf{ALS Evaluation.}
\label{subsection:interpolation}
Our proposed HSR is able to provide a smooth structure to the latent space which is evident by the lower ALS scores of our model. To further analyse the structure of the latent space we perform latent space interpolations and generate a sequence of images. To quantitatively evaluate the interpolation results, we used the proposed ALS scores for the interpolations. The lower ALS score represent the latent space is well structured and the magnitude of the attributes are linearly correlated with the latent transformation. The ALS scores for our model and baseline model in Table. ~\ref{tab:attribute_ppl_score} for following set of popular attributes \{gender, smile, age, hair, bangs, beard\}~\cite{shen2020interfacegan,shen2021closedform,harkonen2020ganspace}. Additionally, Fig.~\ref{tab:attribute_ppl_score} (right) shows the variation of the mean attribute delta ($\Delta_{t,\cdot}$) with the interpolation parameter $t$. We can observe that in the middle region $t\in{[0.4,0.8]}$ the baseline model has high deviation from linear behaviour, which is significantly less in our HSR regularized model. This is also seen quantitatively through proposed ALS-attribute score, in which our model outperforms baseline by \textbf{15\%} of relative improvement. 
We can observe that the interpolations generated using the HSR results in smooth transitions and has high visual quality throughout the interpolation. The StyleGAN2+ADA model without HSR has sudden transitions in between and has some artifacts present (ref. Fig.~\ref{fig:teaser}).

\setlength{\intextsep}{0pt}
\begin{figure*}[!ht]
    \centering
    \includegraphics[width=\linewidth]{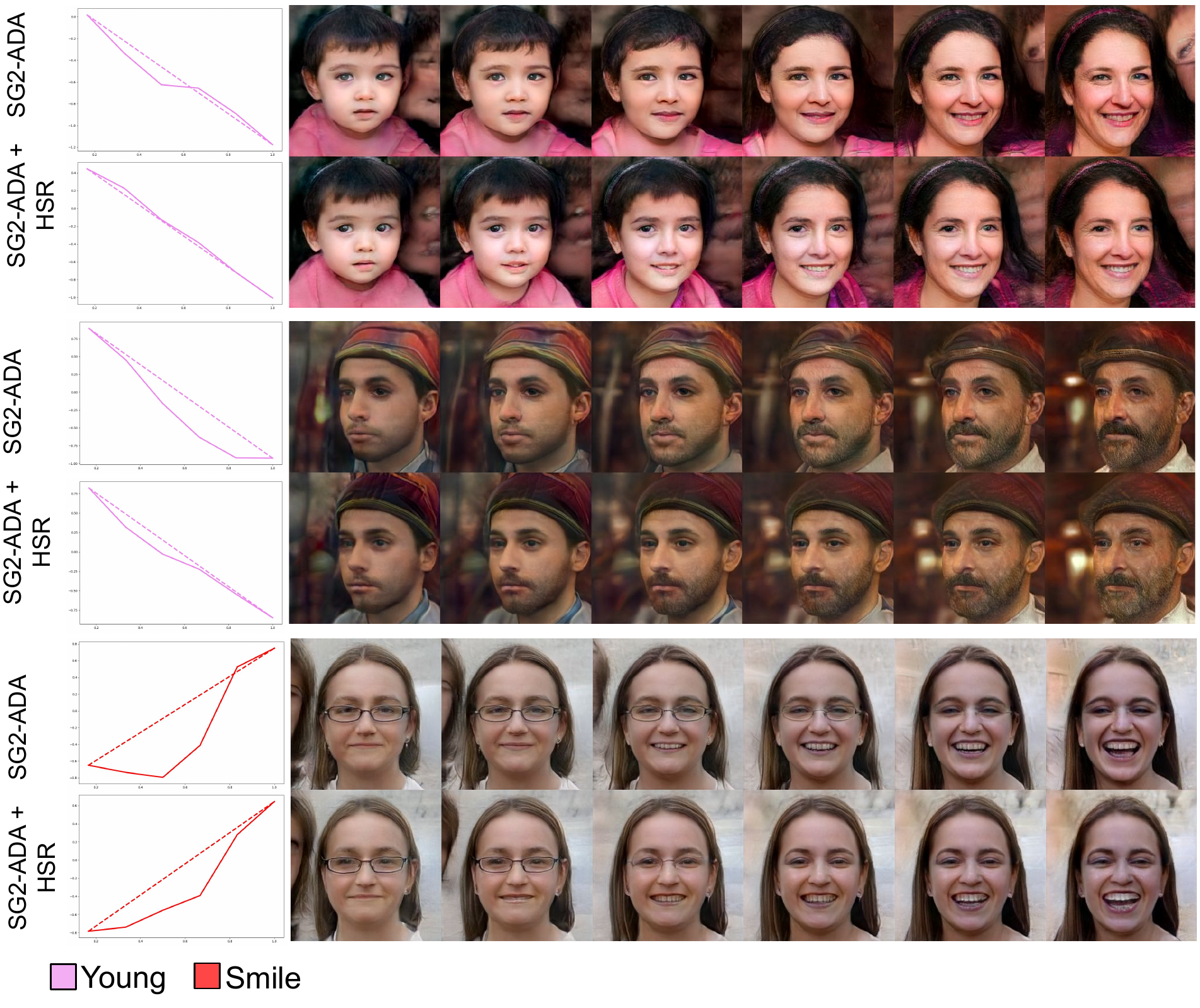}
    \caption{\textbf{Applying HSR improves the linearity of change in attributes.} Here we show improved linearity for ``Young" and ``Smile" attributes. Plots show attribute score on Y-axis, interpolation variable \textit{t} on X-axis.}
    \label{fig:attr-edit-comparison}
\end{figure*}

\noindent \textbf{Editability.} The semantically rich structure of the latent space is widely used for performing semantic edits on the generated images~\cite{shen2020interfacegan,styleflow,wu2021stylespace,patashnik2021styleclip,latentclr,samage}. 
For instance, if we have to add the attribute smile to the generated face image, one can edit the latent code as $\mathbf{w_{edit}} = \mathbf{w} + \alpha\mathbf{d}$ where $\alpha$ is edit strength and $\mathbf{d}$ is the direction for the smile attribute edit operation.
However, often, the attribute scores of the edits performed by such methods does not change linearly with the edit strength parameter $\alpha$ as observed in Fig.~\ref{fig:attr-edit-comparison}. To this end, we perform the following experiment: Given an input source image $I_0$, we first perform attribute edit on it using latent space transformation to obtain $I_1$ using an existing approach~\cite{shen2020interfacegan}. Then, we use the latent code optimization to find the corresponding latent codes $\mathbf{w_0}$ and $\mathbf{w_1}$ in the latent space. Finally, we followed the same approach explained in Sec.~\ref{subsection:interpolation} to generate intermediate images $I_t$ using $w_t$ for $t \{ \in 0, \frac{1}{N}, \frac{2}{N}, ...  1\}$. The results of the interpolation for edits are shown in Fig. ~\ref{fig:attr-edit-comparison}. We compared StyleGAN2-ADA with and without HSR in this experiment. One can observe that in all the cases, adding HSR resulted in added linearity in the attribute scores plots. This property is highly desired in editing methods as it provides a fine-grained control over the attributes in the generated images. Also, observe that both the models evaluated are following the linear line closely in the first two examples. This suggests that the transitions along the age attribute is much more interpretable as it follows linearity. 
In all the three examples the model with HSR is able to approximate the linear line the better than the baseline without HSR. From the images, we can visually observe that the interpolations produced smooth transitions and their is no sudden jump in the attribute when using HSR.  Also note that, the first and last images from both the models do not match ``pixel perfectly", as they is generated by optimization of latent code by different models (with and without HSR). Note that HSR improves the reconstruction quality of real images when embedded in the latent space using projection (ref. supl. mat.). 
\section{Conclusion}
We proposed a novel, hierarchical semantic regularizer called HSR which allows us to regularize the latent representations in StyleGANs by aligning them to semantically rich ones learnt by state-of-the-art classifiers trained on large datasets. HSR is shown to significantly improve the quality of the generated images, especially those created via linear interpolation between attributes corresponding to real images. It further has a desirable property that the latent attribute space becomes more linear. To measure linearity, a novel metric \textit{Attribute Linearity Score (ALS)} was introduced. Copious experiments on standard benchmarks validate the benefits of HSR and  demonstrate statistically significant improvement in the quality of synthesized images. This leads us to interesting avenues for the future work: Enforcing structural priors (\eg linear) in the latent space while training a GAN, which can lead to easier and fine-grained attribute editing.


\noindent \textbf{Acknowledgements.} This work was supported by MeitY (Ministry of Electronics and Information Technology) project (No. 4(16)2019-ITEA), Govt. of India and Uchhatar Avishkar Yojana (UAY) project (IISC\_010), MHRD, India.


\clearpage

\appendix

\renewcommand \thepart{}
\renewcommand \partname{}

\noindent
\begin{center}
\textbf{\Large Supplementary Material: \\ Hierarchical Semantic Regularization of Latent Spaces in StyleGANs} 
\end{center}
\renewcommand{\labelitemii}{$\circ$}

This supplementary document is organized as follows:
\begin{itemize}
\setlength{\itemindent}{-0mm}
    \item Section~\ref{sup:sec:impl}: Implementation Details
    \item Section~\ref{sup:sec:quant}: Additional Quantitative Results 
    \begin{itemize}
        \setlength{\itemindent}{-0mm}
        \item LSUN-Church (Sec.\ \ref{sup:subsec:church})
        \item Additional Metrics for Latent Space Evaluation (Sec.~\ref{sup:subsec:latenteval})
        \item Evaluation on Real Images (Sec.~\ref{sup:sec:realimg})
    \end{itemize}
    \item Section~\ref{sup:sec:qual_results}: Qualitative results
    \begin{itemize}
        \setlength{\itemindent}{-0mm}
        \item Improvement in Worst Images (Sec.~\ref{sup:subsec:worst_imgs})
        \item Linearity of Edits (Sec.~\ref{sup:subsec:lin_edits})
    \end{itemize}
    \item Section~\ref{sup:sec:shortcomings}: Shortcomings of ALS
\end{itemize}{}

\section{Implementation Details}
\label{sup:sec:impl}
Our implementation utilized the  StyleGAN2-ADA~\cite{Karras2020ada} code. We perform all our experiments on 256$\times$256 resolution images. Therefore, we use \texttt{paper256} architecture of StyleGAN2-ADA in all our experiments. 
\begin{figure}
    \centering
    \includegraphics[width=0.99\linewidth]{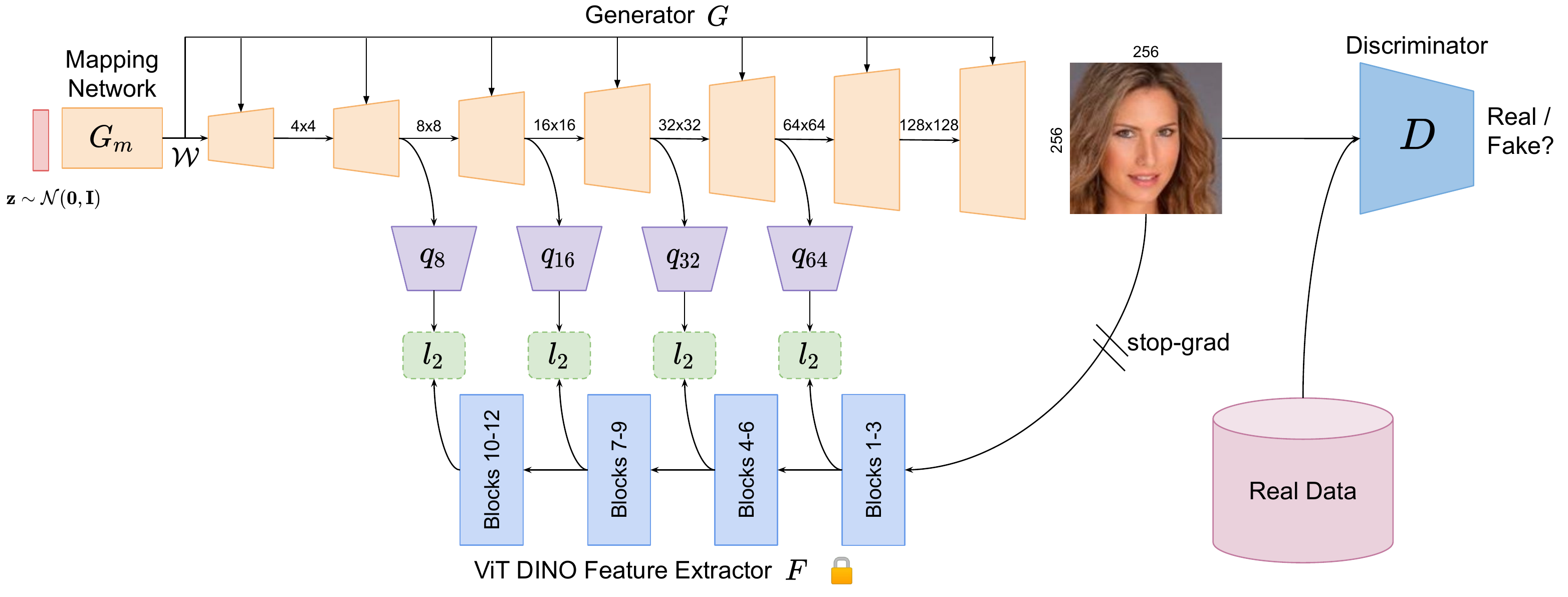}
    \caption{Architectural Overview}
    \label{fig:supp:approach}
\end{figure}

\vspace{1mm} \noindent \textbf{Warm-up. }In order to leverage the rich feature space of pretrained networks using the generated images, we turn on the HSR regularizer after training the GAN for 500kimgs. By this point in the training, the GAN learns to generate images that start to look like the real images.  

\vspace{1mm} \noindent\textbf{HSR. }We use the feature extractor of ViT-DINO~\cite{Caron_2021_ICCV} to extract features. We resize the image to 224$\times$224 before feeding into ViT-DINO. We use the intermediate output (after disarding CLS token) of its 3, 6, 9, and 12th transformer blocks, to supervise the generator's output at the 64, 32, 16, and 8 resolution respectively to align the semantics at various hierarchical levels, since ViT-DINO has been to shown to have a high-to-low level semantics emerging in its stack of transformer blocks ~\cite{vitdino_seg}. We resize the generator's intermediate outputs to 14$\times$14 before applying the loss function ($l_2$).

\section{Additional Quantitative Results}
\label{sup:sec:quant}
\subsection{LSUN-Church}
\label{sup:subsec:church}
\begin{table}
\centering
\caption{Applying the HSR metric yields significant improvements over StyleGAN2 on quality as well as diversity metrics on the LSUN-Church dataset.}
\label{tab:church}
\begin{tabular}{lllll}
\toprule
LSUN-Church & {FID} & {Precision} & {Recall} & {PPL} \\ \midrule
StyleGAN2   & 4.08 & \textbf{0.60} & 0.34                   & 916.15                  \\
+ HSR       & \textbf{3.82} & \textbf{0.60} & \textbf{0.41} & \textbf{678.55}         \\ \bottomrule
\end{tabular}
\end{table}
In addition to the FFHQ-140k results presented in the main paper, we also perform experiment of adding HSR to StyleGAN2 training on LSUN-Church~\cite{yu15lsun} which contains 126k images. We present the quantitative results of this experiment in Table~\ref{tab:church}. We observe a 25.93\% improvement in the PPL over the baseline. Additionally, we also observe a significant boost in the recall metric~\cite{improvedprecisionrecall}, indicating increased diversity in the generated images. 
\subsection{Additional Metrics for Latent Space Evaluation}
\label{sup:subsec:latenteval}

\begin{table}[]
\centering
\caption{Improved Disentanglement}
\label{tab:dci}
\begin{tabular}{llll}
\toprule
               & Disentanglement & Completeness & Informativeness \\ \midrule
SG2-ADA     & 0.57            & 0.61         & 0.98            \\
+HSR & \textbf{0.61}            & \textbf{0.62}         & 0.98            \\  \bottomrule
\end{tabular}
\end{table}

To measure the quality of the $\mathcal{W}$ space after applying HSR, we use the DCI metric~\cite{eastwood2018a}. DCI stands for disentanglement, completeness, and informativeness. Disentanglement measures the extent to which each dimension in the latent captures at most one attribute. Completeness measures the extent of each attribute is controlled by at most one latent dimension. Informativeness measures the classification accuracy of the attributes using latent representation. We use the procedure same as Wu \etal ~\cite{wu2021stylespace} to compute DCI on $\mathcal{W}$ space. We present the results in Table~\ref{tab:dci}, where it is observed that applying HSR improves the disentanglement in $\mathcal{W}$ space while also showing marginal gains in the completeness metric.

\subsection{Evaluation on Real Images}
\label{sup:sec:realimg}
\begin{figure}
    \centering
    \includegraphics[width=0.5\linewidth]{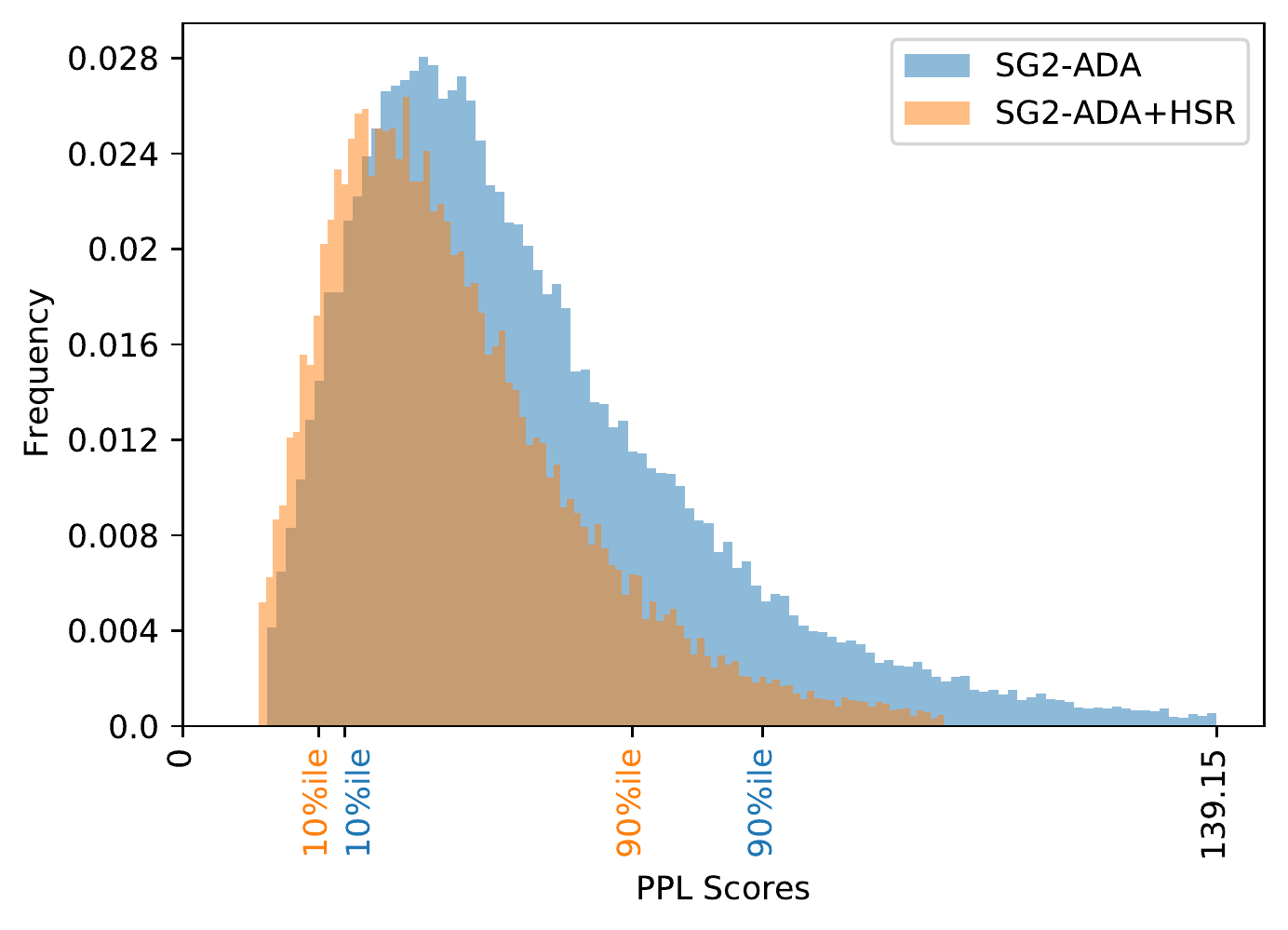}
    \caption{Distribution of PPL scores over 50k real image pairs from CelebA-HQ~\cite{liu2015faceattributes}. Baseline: 46.29, Baseline+HSR: 36.80}
    \label{fig:supp:ppl_real}
\end{figure}
\vspace{1mm} \noindent \textbf{Smoothness of Latent Space.}
In the main paper, we presented results of PPL over 50k generated image pairs. Generated images can also include qualitatively worse images from the learnt distribution, which elongates the tail in PPL distribution. To show PPL on perceptually better images, we use 50k pairs of real images and find their latents. We project 400 randomly sampled images from CelebA to obtain its latents in $\mathcal{W+}$ space. We then randomly sample pairs of inverted latents to compute PPL. We visualize the distribution of PPL scores in Fig.~\ref{fig:supp:ppl_real}. Since real face images are used, the PPL is low as there are no out-of-distribution images (non-face, unnatural face, artefacts). Yet, it can be seen that PPL is lesser when HSR is applied (36.80), compared to the baseline which leads to larger average distance to traversed (46.29) while interpolating between 2 real images.

\vspace{1mm} \noindent \textbf{Reconstruction of Real Images.}
Table~\ref{tab:inv} presents the effectiveness of the StyleGANs to reconstruct/invert real images without and with the application of HSR. Reconstruction, as measured by PSNR and LPIPS metrics, improves when HSR is applied, thus showing that the latent space obtained after regularizing by the HSR leads to more natural-looking images.

\begin{table}
\caption{Effect of HSR on reconstruction of inverted real images}
\label{tab:inv}
\centering
\begin{tabular}{l|ll}
\toprule
Method & PSNR ($\uparrow$) & LPIPS ($\downarrow$) \\ \midrule
SG2-ADA & 47.04 & 0.2296 \\
SG2-ADA+HSR & \textbf{47.16} & \textbf{0.2281} \\
\bottomrule
\end{tabular}
\end{table}

\subsection{Effect of HSR under Distribution of High Quality Images}

Table ~\ref{tab:trunc} presents the results under the truncation trick~\cite{Marchesi2017MegapixelSI}, which is used to produce higher quality images. This comparison, using the FID, Precision-Recall, and PPL metrics in rows 3,4, shows that HSR leads to performance gains even on such higher quality generations. Truncation trades off diversity (recall) for increased quality (precision), irrespective of the use of HSR. On the other hand, using HSR on SG2 (or SG2-Trunc) significantly improves diversity (recall) and FID without sacrificing the quality (precision). Truncation on StyleGAN2, trained on FFHQ140k data, is performed using the commonly used truncation value of $\Psi=0.7$.

\begin{table}[!htbp]
\caption{Comparison of quality and diversity metrics under Truncation}
\label{tab:trunc}
\centering
\begin{tabular}{l|llll}
\toprule
Method        & FID ($\downarrow$)           & Precision ($\uparrow$)    & Recall ($\uparrow$)       & PPL ($\downarrow$)\\ \midrule
SG2             & {3.92}          & \textbf{0.68} & {0.45}          & {175.09}          \\
\rowcolor[HTML]{E2E2E2} SG2+HSR    & \textbf{3.74} & \textbf{0.68} & \textbf{0.48} & \textbf{144.59} \\ \hline
SG2-Trunc   & 21.46         & {\underline{0.83}}    & 0.22          & 109.96 \\
\rowcolor[HTML]{E2E2E2} 
SG2+HSR-Trunc & {\underline{16.94}}   & 0.82          & {\underline{0.25}}    & {\underline {96.46}}\\ \bottomrule
\end{tabular}
\end{table}

\section{Qualitative results}
\label{sup:sec:qual_results}
\subsection{Improvement in Worst Images}
\label{sup:subsec:worst_imgs}

\begin{figure}
    \centering
    \includegraphics[width=0.99\linewidth]{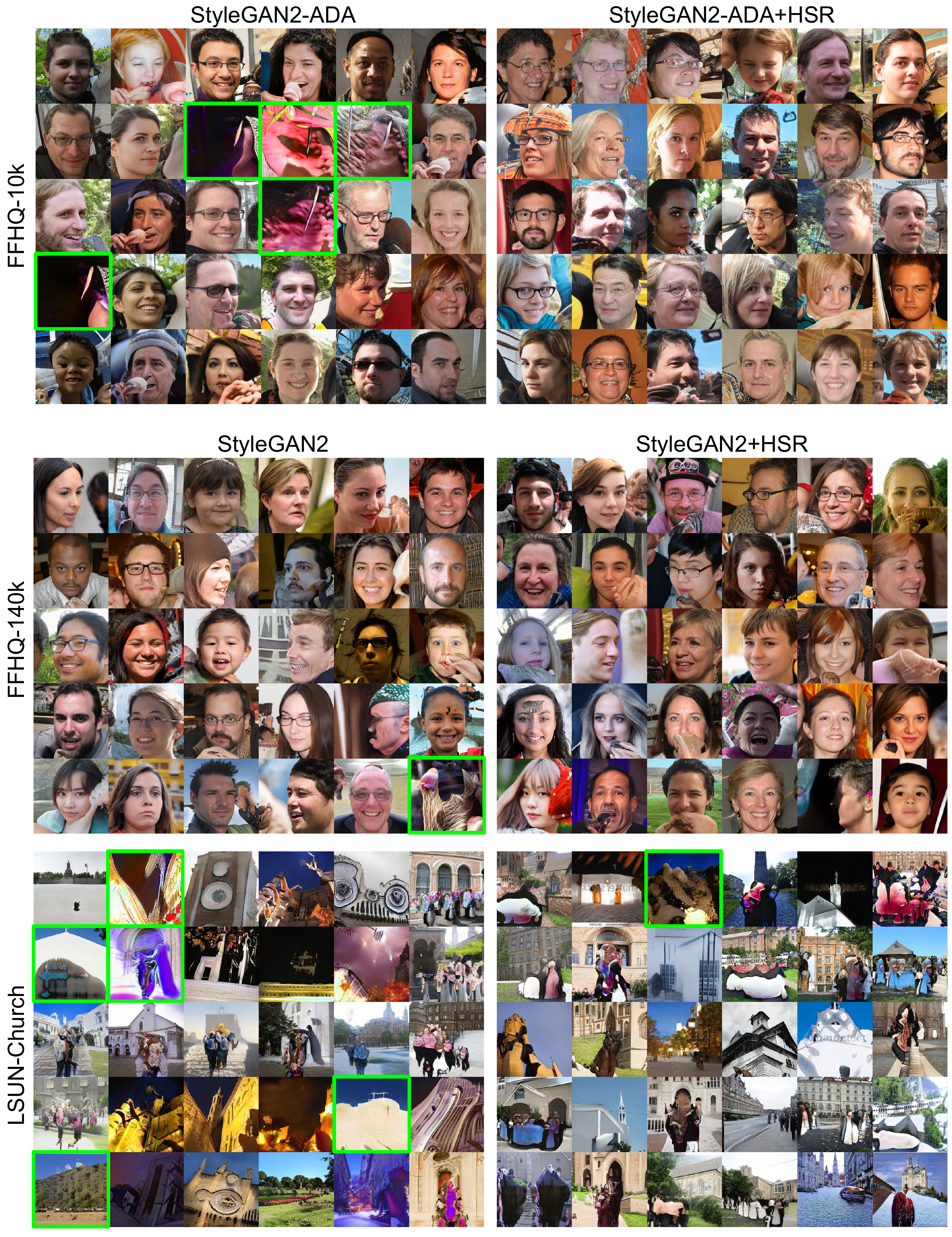}
    \caption{Worst 30 Images according to the Mahalanobis distance to Inception moments of respective datasets. Highlighted images show structural irregularities in the respective image category (face/church). }
    \label{fig:supp:worst_fid}
\end{figure}

We have shown quantitatively that applying the HSR regularization improves the quality of worst images that the generator can produce. We also demonstrate this qualitatively in 2 ways. First, we compare the Mahalanobis distance between the generated images and the moments of the real data from a set of 5000 generated images. We present the results of 30 farthest images in Fig.~\ref{fig:supp:worst_fid}. It can be seen that unnatural, non-face images are being generated by the baseline, which are virtually absent when HSR is applied. In the case of the LSUN-Church dataset, the images lack in structural aspects related to churches and shows presence of unnatural colors in the image. While after applying HSR, the images reproduce the structure faithfully, for e.g., in the edifices.

\begin{figure}[t]
    \centering
    \includegraphics[width=0.99\linewidth]{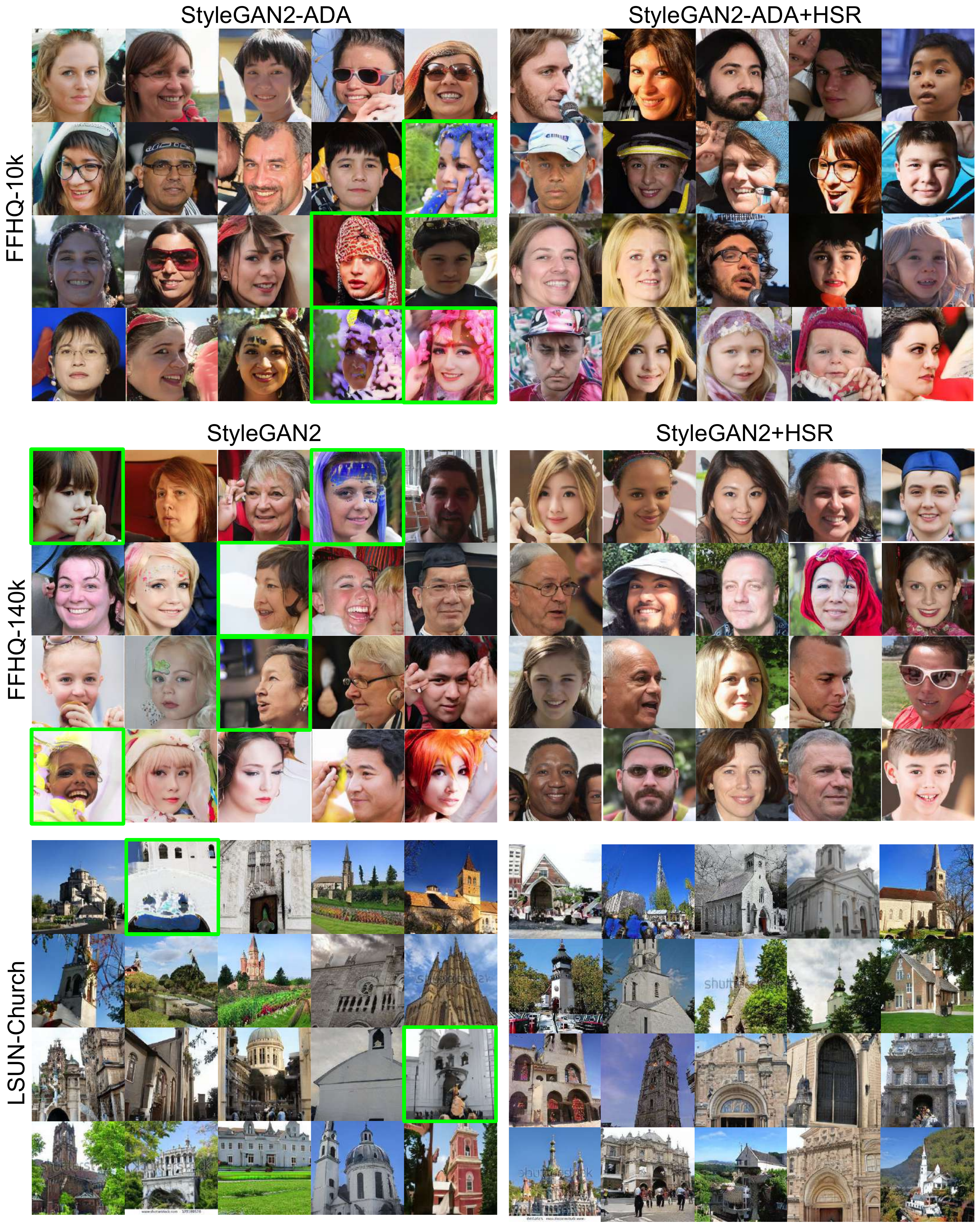}
    \caption{Worst Images according to the PPL scores. Highlighted images have high degree of artefacts.}
    \label{fig:supp:worst_ppl}
\end{figure}

Secondly, we present the results of images sampled from the bottom 10\% according to the  PPL score. We present these results in Fig.~\ref{fig:supp:worst_ppl}. A similar trend is observed with the presence of artefacts in and around the facial regions. In both cases (faces and churches), the artefacts are greatly reduced after the application of the HSR regularizer, making the images look more natural.

\subsection{Linearity of Edits}
\label{sup:subsec:lin_edits}
\begin{figure}
    \centering
    \includegraphics[width=0.85\linewidth]{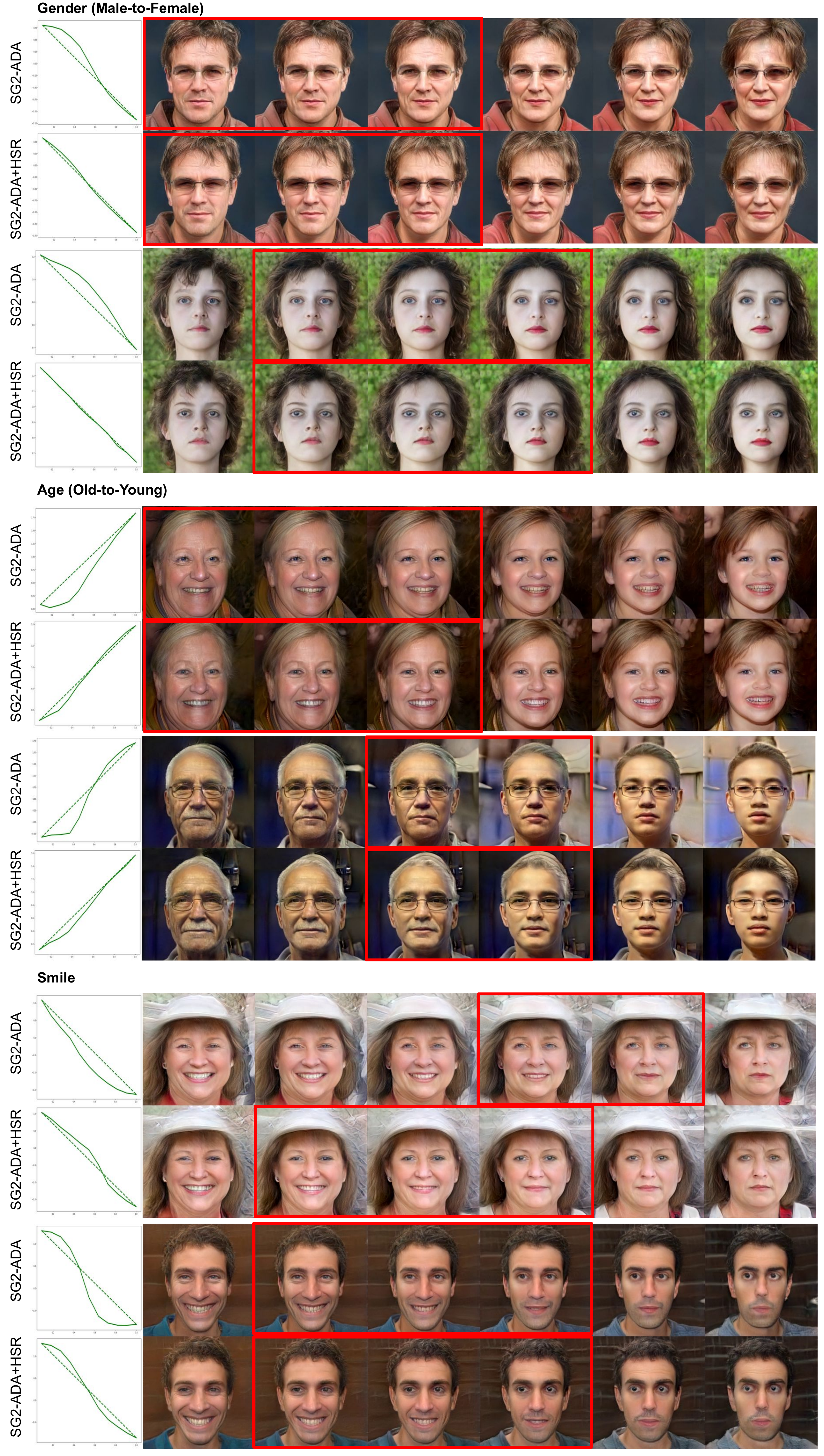}
    \caption{\textbf{Linearity of Edits.} Plots indicate the intensity of attribute . We observe that the rate of change of a particular editable attribute is more close to the identity after applying HSR. In the red inset, we can observe that HSR has added additional smoothness in the transitions in comparison to the SG2-ADA.} 
    \label{fig:supp:edits}
\end{figure}

We present the results of editing along the attributes like ``gender", ``age", and ``smile" in Fig.~\ref{fig:supp:edits}. Note the significant improvement in linearity upon applying the HSR regularizer, as compared to the baseline.

\section{Shortcomings of ALS}
\label{sup:sec:shortcomings}
As noted in the main paper, we propose the ALS score to measure the linearity of change in the attributes. This requires the attributes should not be binary/ categorical but can be continuously varied. Common attributes like age, smile, gender, hair, beard, and bangs fit this description and hence they are a natural choice for evaluation using the ALS metric. Other attributes may be binary or categorical but they still may allow us to evaluate diversity. This is the case for attributes corresponding to ``wearables", for \eg eyeglasses, earrings, headgear etc. The quality of corresponding latent space w.r.t. this class of attributes is better measured with diversity-measuring metrics like recall~\cite{improvedprecisionrecall}.

\clearpage 







\bibliographystyle{splncs04}
\bibliography{egbib}
\end{document}